# What is the Minimal Set of Fragments that Achieves Maximal Parse Accuracy?

**Rens Bod**

School of Computing
University of Leeds, Leeds LS2 9JT, &
Institute for Logic, Language and Computation
University of Amsterdam, Spuistraat 134, 1012 VB Amsterdam
rens@comp.leeds.ac.uk

## Abstract

We aim at finding the minimal set of fragments which achieves maximal parse accuracy in Data Oriented Parsing. Experiments with the Penn Wall Street Journal treebank show that counts of almost arbitrary fragments within parse trees are important, leading to improved parse accuracy over previous models tested on this treebank (a precision of 90.8% and a recall of 90.6%). We isolate some dependency relations which previous models neglect but which contribute to higher parse accuracy.

## 1 Introduction

One of the goals in statistical natural language parsing is to find the minimal set of statistical dependencies (between words and syntactic structures) that achieves maximal parse accuracy. Many stochastic parsing models use linguistic intuitions to find this minimal set, for example by restricting the statistical dependencies to the locality of headwords of constituents (Collins 1997, 1999; Eisner 1997), leaving it as an open question whether there exist important statistical dependencies that go beyond linguistically motivated dependencies. The Data Oriented Parsing (DOP) model, on the other hand, takes a rather extreme view on this issue: given an annotated corpus, all fragments (i.e. subtrees) seen in that corpus, regardless of size and lexicalization, are in principle taken to form a grammar (see Bod 1993, 1998; Goodman 1998; Sima'an 1999). The set of subtrees that is used is thus very large and extremely redundant. Both from a theoretical and from a computational perspective we may wonder whether it is possible to impose constraints on the subtrees that are used, in such a way that the accuracy of the model does not deteriorate or perhaps even improves. That is the main question addressed in this paper. We report on experiments carried out with the Penn Wall Street Journal (WSJ) treebank to investigate several strategies for constraining the set of subtrees. We found that the only constraints that do not decrease the parse accuracy consist in an upper bound of the number of words in the subtree frontiers and an upper bound on the depth of unlexicalized subtrees. We also found that counts of subtrees with several nonheadwords are important, resulting in improved parse accuracy over previous parsers tested on the WSJ.

## 2 The DOP1 Model

To-date, the Data Oriented Parsing model has mainly been applied to corpora of trees whose labels consist of primitive symbols (but see Bod & Kaplan 1998; Bod 2000c, 2001). Let us illustrate the original DOP model presented in Bod (1993), called DOP1, with a simple example. Assume a corpus consisting of only two trees:

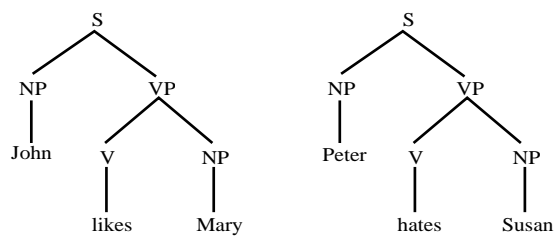

Figure 1. A corpus of two trees

New sentences may be derived by combining fragments, i.e. subtrees, from this corpus, by means of a node-substitution operation indicated as ○. Node-substitution identifies the leftmost nonterminal frontier node of one subtree with the root node of a second subtree (i.e., the second subtree is *substituted* on the leftmost nonterminal frontier node of the first subtree). Thus a new sentence such as *Mary likes Susan* can be derived by combining subtrees from this corpus:

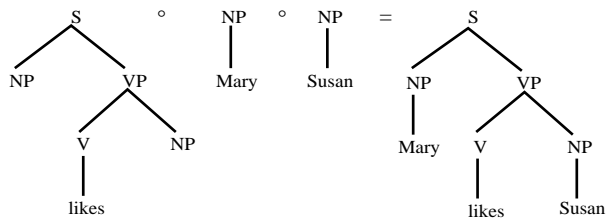

Figure 2. A derivation for *Mary likes Susan*

Other derivations may yield the same tree, e.g.:

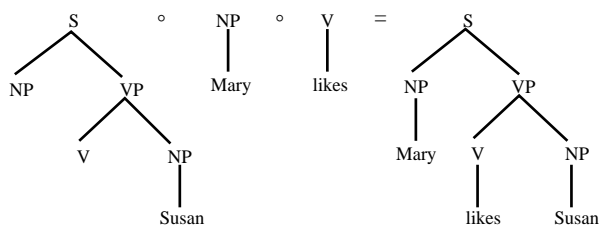

Figure 3. Another derivation yielding same tree

DOP1 computes the probability of a subtree $t$ as the probability of selecting $t$ among all corpus subtrees that can be substituted on the same node as $t$. This probability is equal to the number of occurrences of $t$, $|t|$, divided by the total number of occurrences of all subtrees $t'$ with the same root label as $t$. Let $r(t)$ return the root label of $t$. Then we may write:

$$P(t) = \frac{|t|}{\sum_{t': r(t')=r(t)} |t'|}$$

In most applications of DOP1, the subtree probabilities are smoothed by the technique described in Bod (1996) which is based on Good-Turing. (The subtree probabilities are not smoothed by backing off to smaller subtrees, since these are taken into account by the parse tree probability, as we will see.)

The probability of a derivation $t_1°...°t_n$ is computed by the product of the probabilities of its subtrees $t_i$:

$$P(t_1°...°t_n) = \prod_i P(t_i)$$

As we have seen, there may be several distinct derivations that generate the same parse tree. The probability of a parse tree $T$ is thus the sum of the probabilities of its distinct derivations. Let $t_{id}$ be the $i$-th subtree in the derivation $d$ that produces tree $T$, then the probability of $T$ is given by

$$P(T) = \sum_d \prod_i P(t_{id})$$

Thus the DOP1 model considers counts of subtrees of a wide range of sizes in computing the probability of a tree: everything from counts of single-level rules to counts of entire trees. This means that the model is sensitive to the frequency of large subtrees while taking into account the smoothing effects of counts of small subtrees.

Note that the subtree probabilities in DOP1 are directly estimated from their relative frequencies. A number of alternative subtree estimators have been proposed for DOP1 (cf. Bonnema et al 1999), including maximum likelihood estimation (Bod 2000b). But since the relative frequency estimator has so far not been outperformed by any other estimator for DOP1, we will stick to this estimator in the current paper.

## 3 Computational Issues

Bod (1993) showed how standard chart parsing techniques can be applied to DOP1. Each corpus-subtree $t$ is converted into a context-free rule $r$ where the lefthand side of $r$ corresponds to the root label of $t$ and the righthand side of $r$ corresponds to the frontier labels of $t$. Indices link the rules to the original subtrees so as to maintain the subtree's internal structure and probability. These rules are used to create a derivation forest for a sentence (using a CKY parser), and the most probable parse is computed by sampling a sufficiently large number of random derivations from the forest ("Monte Carlo disambiguation", see Bod 1998). While this technique has been successfully applied to parsing the ATIS portion in the Penn Treebank (Marcus et al. 1993), it is extremely time consuming. This is mainly because the number of random derivations that should be sampled to reliably estimate the most probable parse increases exponentially with the sentence length (see Goodman 1998). It is therefore questionable whether Bod's sampling technique can be scaled to larger domains such as the WSJ portion in the Penn Treebank.

Goodman (1996, 1998) showed how DOP1 can be reduced to a compact stochastic context-free grammar (SCFG) which contains exactly eight SCFG rules for each node in the training set trees. Although Goodman's method does still not allow for an efficient computation of the most probable parse (in fact, the problem of computing the most probable parse in DOP1 is NP-hard -- see Sima'an 1999), his method does allow for an efficient computation of the "maximum constituents parse", i.e. the parse tree that is most likely to have the largest number of correct constituents. Goodman has shown on the ATIS corpus that the maximum constituents parse performs at least as well as the most probable parse if all subtrees are used. Unfortunately, Goodman's reduction method is only beneficial if indeed *all* subtrees are used. Sima'an (1999: 108) argues that there may still be an isomorphic SCFG for DOP1 if the corpus-subtrees are restricted in size or lexicalization, but that the number of the rules explodes in that case.

In this paper we will use Bod's subtree-to-rule conversion method for studying the impact of various subtree restrictions on the WSJ corpus. However, we will not use Bod's Monte Carlo sampling technique from complete derivation forests, as this turned out to be prohibitive for WSJ sentences. Instead, we employ a Viterbi *n*-best search using a CKY algorithm and estimate the most probable parse from the 1,000 most probable derivations, summing up the probabilities of derivations that generate the same tree. Although this heuristic does not guarantee that the most probable parse is actually found, it is shown in Bod (2000a) to perform at least as well as the estimation of the most probable parse with Monte Carlo techniques. However, in computing the 1,000 most probable derivations by means of Viterbi it is prohibitive to keep track of all subderivations at each edge in the chart (at least for such a large corpus as the WSJ). As in most other statistical parsing systems we therefore use the pruning technique described in Goodman (1997) and Collins (1999: 263-264) which assigns a score to each item in the chart equal to the product of the inside probability of the item and its prior probability. Any item with a score less than $10^{-5}$ times of that of the best item is pruned from the chart.

## 4 What is the Minimal Subtree Set that Achieves Maximal Parse Accuracy?

### 4.1 The base line

For our base line parse accuracy, we used the now standard division of the WSJ (see Collins 1997, 1999; Charniak 1997, 2000; Ratnaparkhi 1999) with sections 2 through 21 for training (approx. 40,000 sentences) and section 23 for testing (2416 sentences ≤ 100 words); section 22 was used as development set. All trees were stripped off their semantic tags, co-reference information and quotation marks. We used all training set subtrees of depth 1, but due to memory limitations we used a subset of the subtrees larger than depth 1, by taking for each depth a random sample of 400,000 subtrees. These random subtree samples were not selected by first exhaustively computing the complete set of subtrees (this was computationally prohibitive). Instead, for each particular depth > 1 we sampled subtrees by randomly selecting a node in a random tree from the training set, after which we selected random expansions from that node until a subtree of the particular depth was obtained. We repeated this procedure 400,000 times for each depth > 1 and ≤ 14. Thus no subtrees of depth > 14 were used. This resulted in a *base line subtree set* of 5,217,529 subtrees which were smoothed by the technique described in Bod (1996) based on Good-Turing. Since our subtrees are allowed to be lexicalized (at their frontiers), we did not use a separate part-of-speech tagger: the test sentences were directly parsed by the training set subtrees. For words that were unknown in our subtree set, we guessed their categories by means of the method described in Weischedel et al. (1993) which uses statistics on word-endings, hyphenation and capitalization. The guessed category for each unknown word was converted into a depth-1 subtree and assigned a probability by means of simple Good-Turing estimation (see Bod 1998). The most probable parse for each test sentence was estimated from the 1,000 most probable derivations of that sentence, as described in section 3.

We used "evalb"[1] to compute the standard PARSEVAL scores for our parse results. We focus on the Labeled Precision (LP) and Labeled Recall (LR) scores only in this paper, as these are commonly used to rank parsing systems.

Table 1 shows the LP and LR scores obtained with our base line subtree set, and compares these scores with those of previous stochastic parsers tested on the WSJ (respectively Charniak 1997, Collins 1999, Ratnaparkhi 1999, and Charniak 2000).

The table shows that by using the base line subtree set, our parser outperforms most previous parsers but it performs worse than the parser in Charniak (2000). We will use our scores of 89.5% LP and 89.3% LR (for test sentences ≤ 40 words) as the base line result against which the effect of various subtree restrictions is investigated. While most subtree restrictions diminish the accuracy scores, we will see that there are restrictions that improve our scores, even beyond those of Charniak (2000).

---

[1] http://www.cs.nyu.edu/cs/projects/proteus/evalb/

We will initially study our subtree restrictions only for test sentences ≤ 40 words (2245 sentences), after which we will give in 4.6 our results for all test sentences ≤ 100 words (2416 sentences). While we have tested all subtree restrictions initially on the development set (section 22 in the WSJ), we believe that it is interesting and instructive to report these subtree restrictions on the test set (section 23) rather than reporting our best result only.

| Parser | LP | LR |
|---|---|---|
| ≤ 40 words | | |
| Char97 | 87.4 | 87.5 |
| Coll99 | 88.7 | 88.5 |
| Char00 | 90.1 | 90.1 |
| Bod00 | 89.5 | 89.3 |
| ≤ 100 words | | |
| Char97 | 86.6 | 86.7 |
| Coll99 | 88.3 | 88.1 |
| Ratna99 | 87.5 | 86.3 |
| Char00 | 89.5 | 89.6 |
| Bod00 | 88.6 | 88.3 |

Table 1. Parsing results with the base line subtree set compared to previous parsers

### 4.2 The impact of subtree size

Our first subtree restriction is concerned with subtree size. We therefore performed experiments with versions of DOP1 where the base line subtree set is restricted to subtrees with a certain maximum depth. Table 2 shows the results of these experiments.

| depth of subtrees | LP | LR |
|---|---|---|
| 1 | 76.0 | 71.8 |
| ≤2 | 80.1 | 76.5 |
| ≤3 | 82.8 | 80.9 |
| ≤4 | 84.7 | 84.1 |
| ≤5 | 85.5 | 84.9 |
| ≤6 | 86.2 | 86.0 |
| ≤8 | 87.9 | 87.1 |
| ≤10 | 88.6 | 88.0 |
| ≤12 | 89.1 | 88.8 |
| ≤14 | 89.5 | 89.3 |

Table 2. Parsing results for different subtree depths (for test sentences ≤ 40 words)

Our scores for subtree-depth 1 are comparable to Charniak's treebank grammar if tested on word strings (see Charniak 1997). Our scores are slightly better, which may be due to the use of a different unknown word model. Note that the scores consistently improve if larger subtrees are taken into account. The highest scores are obtained if the full base line subtree set is used, but they remain behind the results of Charniak (2000). One might expect that our results further increase if even larger subtrees are used; but due to memory limitations we did not perform experiments with subtrees larger than depth 14.

### 4.3 The impact of lexical context

The more words a subtree contains in its frontier, the more lexical dependencies can be taken into account. To test the impact of the lexical context on the accuracy, we performed experiments with different versions of the model where the base line subtree set is restricted to subtrees whose frontiers contain a certain maximum number of words; the subtree depth in the base line subtree set was not constrained (though no subtrees deeper than 14 were in this base line set). Table 3 shows the results of our experiments.

| # words in subtrees | LP | LR |
|---|---|---|
| ≤1 | 84.4 | 84.0 |
| ≤2 | 85.2 | 84.9 |
| ≤3 | 86.6 | 86.3 |
| ≤4 | 87.6 | 87.4 |
| ≤6 | 88.0 | 87.9 |
| ≤8 | 89.2 | 89.1 |
| ≤10 | 90.2 | 90.1 |
| ≤11 | 90.8 | 90.4 |
| ≤12 | 90.8 | 90.5 |
| ≤13 | 90.4 | 90.3 |
| ≤14 | 90.3 | 90.3 |
| ≤16 | 89.9 | 89.8 |
| unrestricted | 89.5 | 89.3 |

Table 3. Parsing results for different subtree lexicalizations (for test sentences ≤ 40 words)

We see that the accuracy initially increases when the lexical context is enlarged, but that the accuracy decreases if the number of words in the subtree frontiers exceeds 12 words. Our highest scores of 90.8% LP and 90.5% LR outperform the scores of the best previously published parser by Charniak (2000) who obtains 90.1% for both LP and LR. Moreover, our scores also outperform the reranking technique of Collins (2000) who reranks the output of the parser of Collins (1999) using a boosting method based on Schapire & Singer (1998), obtaining 90.4% LP and 90.1% LR. We have thus found a subtree restriction which does not decrease the parse accuracy but even improves it. This restriction consists of an upper bound of 12 words in the subtree frontiers, for subtrees ≤ depth 14. (We have also tested this lexical restriction in combination with subtrees smaller than depth 14, but this led to a decrease in accuracy.)

### 4.4 The impact of structural context

Instead of investigating the impact of lexical context, we may also be interested in studying the importance of structural context. We may raise the question as to whether we need all *un*lexicalized subtrees, since such subtrees do not contain any lexical information, although they may be useful to smooth lexicalized subtrees. We accomplished a set of experiments where unlexicalized subtrees of a certain minimal depth are deleted from the base line subtree set, while all lexicalized subtrees up to 12 words are retained.

| depth of deleted unlexicalized subtrees | LP | LR |
|---|---|---|
| ≥1 | 79.9 | 77.7 |
| ≥2 | 86.4 | 86.1 |
| ≥3 | 89.9 | 89.5 |
| ≥4 | 90.6 | 90.2 |
| ≥5 | 90.7 | 90.6 |
| ≥6 | 90.8 | 90.6 |
| ≥7 | 90.8 | 90.5 |
| ≥8 | 90.8 | 90.5 |
| ≥10 | 90.8 | 90.5 |
| ≥12 | 90.8 | 90.5 |

Table 4. Parsing results for different structural context (for test sentences ≤ 40 words)

Table 4 shows that the accuracy increases if unlexicalized subtrees are retained, but that unlexicalized subtrees larger than depth 6 do not contribute to any further increase in accuracy. On the contrary, these larger subtrees even slightly decrease the accuracy. The highest scores obtained are: 90.8% labeled precision and 90.6% labeled recall. We thus conclude that pure structural context without any lexical information contributes to higher parse accuracy (even if there exists an upper bound for the size of structural context). The importance of structural context is consonant with Johnson (1998) who showed that structural context from higher nodes in the tree (i.e. grandparent nodes) contributes to higher parse accuracy. This mirrors our result of the importance of unlexicalized subtrees of depth 2. But our results show that larger structural context (up to depth 6) also contributes to the accuracy.

### 4.5 The impact of nonheadword dependencies

We may also raise the question as to whether we need almost arbitrarily large *lexicalized* subtrees (up to 12 words) to obtain our best results. It could be the case that DOP's gain in parse accuracy with increasing subtree depth is due to the model becoming sensitive to the influence of lexical heads higher in the tree, and that this gain could also be achieved by a more compact model which associates each nonterminal with its headword, such as a head-lexicalized SCFG. Head-lexicalized stochastic grammars have recently become increasingly popular (see Collins 1997, 1999; Charniak 1997, 2000). These grammars are based on Magerman's head-percolation scheme to determine the headword of each nonterminal (Magerman 1995). Unfortunately this means that head-lexicalized stochastic grammars are not able to capture dependency relations between words that according to Magerman's head-percolation scheme are "nonheadwords" -- e.g. between *more* and *than* in the WSJ construction *carry more people than cargo* where neither *more* nor *than* are headwords of the NP constituent *more people than cargo*. A frontier-lexicalized DOP model, on the other hand, captures these dependencies since it includes subtrees in which *more* and *than* are the only frontier words. One may object that this example is somewhat far-fetched, but Chiang (2000) notes that head-lexicalized stochastic grammars fall short in encoding even simple dependency relations such as between *left* and *John* in the sentence *John should have left*. This is because Magerman's head-percolation scheme makes *should* and *have* the heads of their respective VPs so that there is no dependency relation between the verb *left* and its subject *John*. Chiang observes that almost a quarter of all nonempty subjects in the WSJ appear in such a configuration.

In order to isolate the contribution of nonheadword dependencies to the parse accuracy, we eliminated all subtrees containing a certain maximum number of nonheadwords, where a nonheadword of a subtree is a word which according to Magerman's scheme is not a headword of the subtree's root nonterminal

(although such a nonheadword may of course be a headword of one of the subtree's internal nodes). In the following experiments we used the subtree set for which maximum accuracy was obtained in our previous experiments, i.e. containing all lexicalized subtrees with maximally 12 frontier words and all unlexicalized subtrees up to depth 6.

| # nonheadwords in subtrees | LP | LR |
|---|---|---|
| 0 | 89.6 | 89.6 |
| ≤1 | 90.2 | 90.1 |
| ≤2 | 90.4 | 90.2 |
| ≤3 | 90.3 | 90.2 |
| ≤4 | 90.6 | 90.4 |
| ≤5 | 90.6 | 90.6 |
| ≤6 | 90.6 | 90.5 |
| ≤7 | 90.7 | 90.7 |
| ≤8 | 90.8 | 90.6 |
| unrestricted | 90.8 | 90.6 |

Table 5. Parsing results for different number of nonheadwords (for test sentences ≤ 40 words)

Table 5 shows that nonheadwords contribute to higher parse accuracy: the difference between using no and all nonheadwords is 1.2% in LP and 1.0% in LR. Although this difference is relatively small, it does indicate that nonheadword dependencies should preferably not be discarded in the WSJ. We should note, however, that most other stochastic parsers do include counts of *single* nonheadwords: they appear in the backed-off statistics of these parsers (see Collins 1997, 1999; Charniak 1997; Goodman 1998). But our parser is the first parser that also includes counts between two or more nonheadwords, to the best of our knowledge, and these counts lead to improved performance, as can be seen in table 5.

### 4.6 Results for all sentences

We have seen that for test sentences ≤ 40 words, maximal parse accuracy was obtained by a subtree set which is restricted to subtrees with not more than 12 words and which does not contain unlexicalized subtrees deeper than 6.[2] We used these restrictions to test our model on all sentences ≤ 100 words from the WSJ test set. This resulted in an LP of 89.7% and an LR of 89.7%. These scores slightly outperform the best previously published parser by Charniak (2000), who obtained 89.5% LP and 89.6% LR for test sentences ≤ 100 words. Only the reranking technique proposed by Collins (2000) slightly outperforms our precision score, but not our recall score: 89.9% LP and 89.6% LR.

## 5 Discussion: Converging Approaches

The main goal of this paper was to find the minimal set of fragments which achieves maximal parse accuracy in Data Oriented Parsing. We have found that this minimal set of fragments is very large and extremely redundant: highest parse accuracy is obtained by employing only two constraints on the fragment set: a restriction of the number of words in the fragment frontiers to 12 and a restriction of the depth of unlexicalized fragments to 6. No other constraints were warranted.

There is an important question why maximal parse accuracy occurs with exactly these constraints. Although we do not know the answer to this question, we surmise that these constraints differ from corpus to corpus and are related to general data sparseness effects. In previous experiments with DOP1 on smaller and more restricted domains we found that the parse accuracy decreases also after a certain maximum subtree depth (see Bod 1998; Sima'an 1999). We expect that also for the WSJ the parse accuracy will decrease after a certain depth, although we have not been able to find this depth so far.

A major difference between our approach and most other models tested on the WSJ is that the DOP model uses frontier lexicalization while most other models use constituent lexicalization (in that they associate each constituent nonterminal with its lexical head -- see Collins 1996, 1999; Charniak 1997; Eisner 1997). The results in this paper indicate that frontier lexicalization is a promising alternative to constituent lexicalization. Our results also show that the linguistically motivated constraint which limits the statistical dependencies to the locality of headwords of constituents is too narrow. Not only are counts of subtrees with nonheadwords important, also counts of unlexicalized subtrees up to depth 6 increase the parse accuracy.

The only other model that uses frontier lexicalization and that was tested on the standard WSJ split is Chiang (2000) who extracts a stochastic tree-insertion grammar or STIG (Schabes & Waters 1996) from the WSJ, obtaining 86.6% LP and 86.9% LR for sentences ≤ 40 words. However, Chiang's approach is limited in at least two respects. First, each elementary tree in his STIG is lexicalized with exactly one lexical item, while our results show that there is an increase in parse accuracy if more lexical items and also if unlexicalized trees are included (in his conclusion Chiang acknowledges that "multiply anchored trees" may be important). Second, Chiang computes the probability of a tree by taking into account only one derivation, while in STIG, like in DOP1, there can be several derivations that generate the same tree.

---

[2] It may be noteworthy that for the development set (section 22 of WSJ), maximal parse accuracy was obtained with exactly the same subtree restrictions. As explained in 4.1, we initially tested all restrictions on the development set, but we preferred to report the effects of these restrictions for the test set.

Another difference between our approach and most other models is that the underlying grammar of DOP is based on a treebank grammar (cf. Charniak 1996, 1997), while most current stochastic parsing models use a "markov grammar" (e.g. Collins 1999; Charniak 2000). While a treebank grammar only assigns probabilities to rules or subtrees that are seen in a treebank, a markov grammar assigns probabilities to any possible rule, resulting in a more robust model. We expect that the application of the markov grammar approach to DOP will further improve our results. Research in this direction is already ongoing, though it has been tested for rather limited subtree depths only (see Sima'an 2000).

Although we believe that our main result is to have shown that almost arbitrary fragments within parse trees are important, it is surprising that a relatively simple model like DOP1 outperforms most other stochastic parsers on the WSJ. Yet, to the best of our knowledge, DOP is the only model which does not *a priori* restrict the fragments that are used to compute the most probable parse. Instead, it starts out by taking into account all fragments seen in a treebank and then investigates fragment restrictions to discover the set of relevant fragments. From this perspective, the DOP approach can be seen as striving for the same goal as other approaches but from a different direction. While other approaches usually limit the statistical dependencies beforehand (for example to headword dependencies) and then try to improve parse accuracy by gradually letting in more dependencies, the DOP approach starts out by taking into account as many dependencies as possible and then tries to constrain them without losing parse accuracy. It is not unlikely that these two opposite directions will finally converge to the same, true set of statistical dependencies for natural language parsing.

As it happens, quite some convergence has already taken place. The history of stochastic parsing models shows a consistent increase in the scope of statistical dependencies that are captured by these models. Figure 4 gives a (very) schematic overview of this increase (see Carroll & Weir 2000, for a more detailed account of a subsumption lattice where SCFG is at the bottom and DOP at the top).

| Model | Scope of Statistical Dependencies |
|---|---|
| Charniak (1996) | context-free rules |
| Collins (1996), Eisner (1996) | context-free rules, headwords |
| Charniak (1997) | context-free rules, headwords, grandparent nodes |
| Collins (2000) | context-free rules, headwords, grandparent nodes/rules, bigrams, two-level rules, two-level bigrams, nonheadwords |
| Bod (1992) | all fragments within parse trees |

Figure 4. Schematic overview of the increase of statistical dependencies by stochastic parsers

Thus there seems to be a convergence towards a maximalist model which "takes all fragments [...] and lets the statistics decide" (Bod 1998: 5). While early head-lexicalized grammars restricted the fragments to the locality of headwords (e.g. Collins 1996; Eisner 1996), later models showed the importance of including context from higher nodes in the tree (Charniak 1997; Johnson 1998). This mirrors our result of the utility of (unlexicalized) fragments of depth 2 and larger. The importance of including single nonheadwords is now also uncontroversial (e.g. Collins 1997, 1999; Charniak 2000), and the current paper has shown the importance of including two and more nonheadwords. Recently, Collins (2000) observed that "In an ideal situation we would be able to encode arbitrary features $h_S$, thereby keeping track of counts of arbitrary fragments within parse trees". This is in perfect correspondence with the DOP philosophy.